\documentclass[letterpaper]{article} 
\usepackage{aaai25}  
\usepackage{times}  
\usepackage{helvet}  
\usepackage{courier}  
\usepackage[hyphens]{url}  
\usepackage{graphicx} 
\usepackage{natbib}  
\usepackage{caption} 
\usepackage{algorithm}
\usepackage{multirow}
\usepackage{algpseudocode}
\usepackage{textcomp}
\usepackage{xcolor}
\usepackage{booktabs}
\usepackage{amsthm}
\usepackage{amssymb}
\usepackage{amsmath}
\usepackage{subcaption}
\usepackage{bm}
\newtheorem{definition}{Definition}
\usepackage{newfloat}
\usepackage{listings}
\DeclareCaptionStyle{ruled}{labelfont=normalfont,labelsep=colon,strut=off} 
\floatstyle{ruled}
\newfloat{listing}{tb}{lst}{}
\floatname{listing}{Listing}
\title{Learning-Based Finite Element Methods Modeling for Complex Mechanical Systems}
\author{
    Shi Jiasheng\textsuperscript{\rm 1}, Lin Fu\textsuperscript{\rm 1}, Rao Weixiong\textsuperscript{\rm 1}
}
\affiliations{
    \textsuperscript{\rm 1}School of Software Engineering, Tongji University\\

}

\usepackage{bibentry}

\begin{document}

\maketitle

\begin{abstract}

Complex mechanic systems simulation is important in many real-world applications. The de-facto numeric solver using Finite Element Method (FEM) suffers from computationally intensive overhead. Though with many progress on the reduction of computational time and acceptable accuracy, the recent CNN or GNN-based simulation models still struggle to effectively represent complex mechanic simulation caused by the long-range spatial dependency of distance mesh nodes and independently learning local and global representation. In this paper, we propose a novel two-level mesh graph network. The key of the network is to interweave the developed Graph Block and Attention Block to better learn mechanic interactions even for long-rang spatial dependency. Evaluation on three synthetic and one real datasets demonstrates the superiority of our work. For example, on the Beam dataset, our work leads to 54.3\% lower prediction errors and 9.87\% fewer learnable network parameters.

\end{abstract}

%

\section{Introduction}


Simulation of complex mechanic systems is crucial in many real world applications, e.g., Solid Mechanics \cite{zienkiewicz2000finite} and  Fluid Mechanics \cite{reddy2015introduction}. Partial Differential Equations (PDEs) have been widely used to model the underlying mechanics, and Finite Element Method (FEM) now becomes the de-facto numeric solver for PDEs. It is mainly because FEM simulations provide valuable resources to remove instances of creating and testing expensive product prototypes for high-fidelity situations.

\begin{figure}[t]
\centering\vspace{-1ex}
\includegraphics[width=1\linewidth]{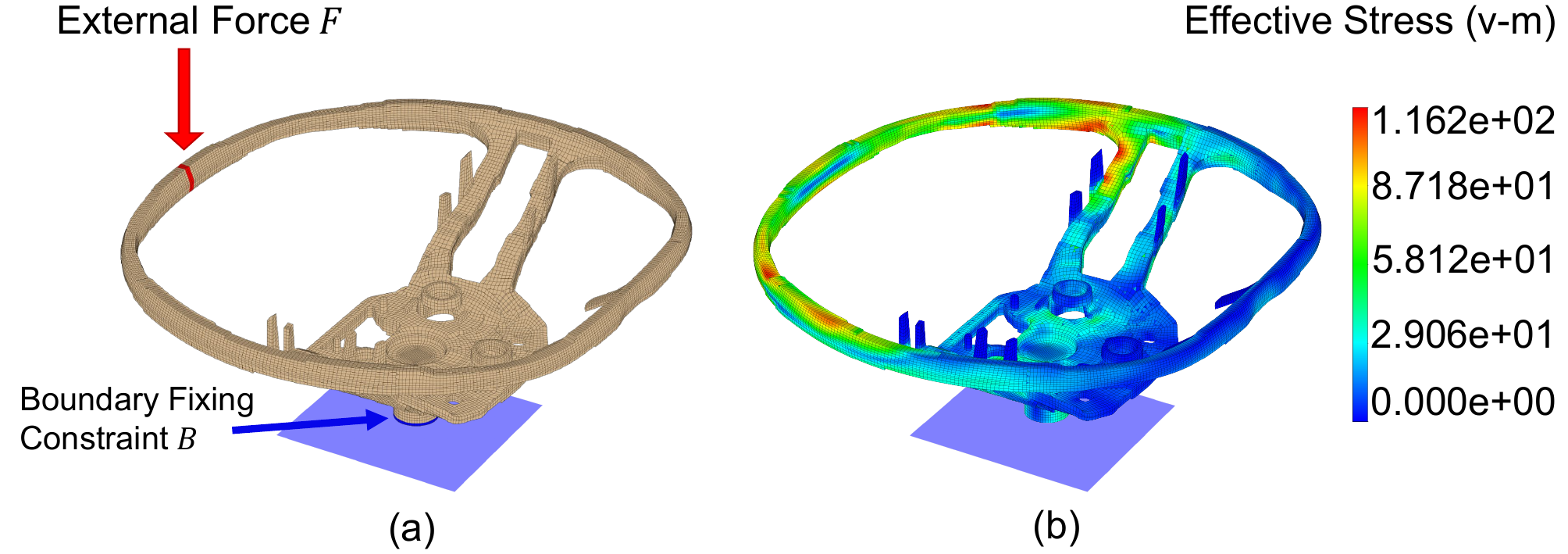}\vspace{-1ex}
\caption{Bending Test of Steering Wheels applied with external force $F$ and fixed on the bottom plane $B$. (a) FEM-divided mesh structure; (b) Stress field simulation result.}\vspace{-2ex}
\label{fig:overview}
\centering
\end{figure}

Fig. \ref{fig:overview} gives the FEM simulation result of an example steering wheel. The left sub-figure shows the mesh structure divided by an FEM mesh generator, and the right one plots the heatmap of \emph{effective stress} on the mesh structure. If the effective stress exceeds a certain threshold, the wheel might twist or even fracture. By FEM simulation, mechanic engineers can easily identify product design defects and then optimize the design for better mechanic performance.

When the number of divided meshes is high (e.g., tens of thousands or even more), solving the PDEs by FEM is computationally intensive and costly. Even with minor changes to a mechanic system, e.g., the force $F$ or the geometry  structure of the wheel, FEM solvers have to recompute the entire simulation, resulting in substantial overhead.

Recently, researchers have explored deep learning techniques to build end-to-end models between simulation inputs and outputs, including Convolutional Neural Networks (CNNs) \cite{DESHPANDE2022115307, NieJK20, thuerey2020deep} and Graph Neural Networks (GNNs) \cite{Sanchez-Gonzalez20, PfaffFSB21, LienenG22, SalehiG22}. Particularly, MeshGraphNet \cite{Sanchez-Gonzalez20, PfaffFSB21} learns mesh graph networks on FEM-divided mesh structures with significant reduction of computational time and acceptable accuracy.

Unfortunately, the learning models still struggle to effectively represent the simulation of complex mechanic systems. The simulation essentially depends upon the \emph{mechanic interactions} between the simulation object (e.g., the wheel), the force $F$ and the fixing constraint, namely boundary condition $B$. As shown in Fig. \ref{fig:overview}, since $F$ and $B$ are applied to a small area of the wheel,  existing works, e.g, MeshGraphNet, via a certain number of GNN message passing steps, may not effectively propagate the response of $F$ from its original small area to an arbitrary mesh graph node, particularly to those within long-range areas. Moreover, learning  mechanic interaction involves both local and global representation, e.g., the force $F$ and constraint $B$ on small areas, and the overall geometry structure of the entire wheel. Effective representation without missing any of them is non-trivial. For example, Eagle \cite{JannyBNDT023} independently learns global embeddings on a coarse level and local ones on the original level before a decoder then concatenates such embeddings to generate mechanic response (stress and velocity). Yet the stress field of the entire wheel (i.e., global representation) heavily depends on the force $F$ on the small area (i.e., local representation), and such independent representation does not make sense.

To tackle the challenges above, in this paper,  by following the Encoder-Processor-Decoder paradigm, we propose a novel two-level mesh graph network. On the fine mesh node graph level, we use the developed Graph Block (GBK) to learn local representation, and next exploit the Attention Block (ABK) to learn global representation on a coarse level. Due to a much smaller number of mesh nodes in  the coarse level than the originally fine level, the ABK block can more efficiently perform the Transformer operation to learn the dependencies between arbitrary mesh nodes. Moreover, instead of independently learning local and global representation, the Processor module involves a sequence of $M$ layers, each of which involves a GBK followed by an ABK. In this way, the Processor sequentially interweaves the ABK and GBK blocks for better local and global representation. As a summary, we make the following contributions.

\begin{itemize}
    \item We propose the two-level mesh graph network to effectively and efficiently learn local and global representation in complex mechanic simulation by the developed ABK and GBK blocks.
    \item We develop the techniques to generate coarse mesh nodes by a simplified Louvain algorithm and encode mesh node spatial positions by a Laplacian encoding scheme. 
    \item Evaluation on three synthetic and one real datasets demonstrates the {superiority} of our work. For example, when  compared to state-of-the-art, on the Beam dataset, our work leads to   $54.30\%$ lower prediction errors and 9.87\% fewer learnable network parameters.
\end{itemize}


\section{Related Works}\label{sec2:relate}

\subsubsection{CNN-based Models} 
When mechanic systems are modeled by regular grids, some works attempt to develop CNN regression models to predict mechanic response. In solid mechanic simulation, the previous work \cite{liang2018deep} developed a CNN model to learn a stress field prediction model by mapping FEM input to the output distribution of aortic wall stress. The work \cite{NieJK20} adopted CNNs to predict the stress field in 2D cantilevered structures with a linear isotropic elastic material subjected to external loads at the free end of such structures. In addition, the works \cite{RaissiPK19, thuerey2020deep} have explored the potential of fluid prediction models with regular grid-like structures in 2D or 3D domains. Nevertheless, these works either explicitly require regular grids or have to pre-process input data into regular grids, such that these works can comfortably build CNN-based simulation models. As a result, it is not hard to find that such works do not perform well in complex mechanic simulation with irregular grids.

\subsubsection{GNN-based Models} For complex mechanic simulation with irregular mesh structures, the works \cite{Sanchez-Gonzalez20, PfaffFSB21} and their follow-up variants \cite{abs-2210-00612,abs-2205-02637,AllenRL0SBP23} proposed flat or hierarchical mesh graph networks to better represent such structures. For example, the previous works \cite{Sanchez-Gonzalez20} employed dynamic particles to represent mechanic systems by mesh graphs, where graph nodes indicate the particles and graph edges are built to connect particles and their proximate neighbours within a certain distance. Next, to simulate rigid collisions among arbitrary shapes, the work \cite{AllenRL0SBP23} introduced the 'Face Interaction Graph Network' (FIGNet), by extending message passing from traditional graphs with directed edges between nodes to proposed graphs with directed hyper-edges between faces.

Unlike the flat mesh graph networks above, some works \cite{abs-2205-02637, DeshpandeBL24} developed hierarchical GNN models with larger receptive field to better represent simulation systems. However, such models do not guarantee a truly global receptive field across the entire system, due to the limited number of GNN message passing steps.

To overcome the issue of limited message passing steps in GNNs, some works \cite{JannyBNDT023, HanGPWL22} have employed Transformers \cite{VaswaniSPUJGKP17} to learn spatial or temporal dependency in mechanic simulations. For example, Eagle \cite{JannyBNDT023} uses Transformers to learn spatial dependency even between long-range graph nodes. Nevertheless, due to independent representation of local and global mechanic interactions, Eagle may not learn the spatial dependency to the best.

\subsubsection{Neural Operator-based Models}
Unlike the end-to-end learning models above, some recent work FNO \cite{LI2020NO} and the improvement Geo-FNO \cite{10.5555/3648699.3649087} propose to replace computing intensive operators within the PDE solving framework by light-weighted neural networks. The key idea of FNO is to employ frequency domain multiplications via Fourier transforms, as an alternative to spatial domain integrals. Yet, FNO is limited to the rectangular domains modeled by uniform grids. Geo-FNO overcomes this issue by introducing learnable deformation from irregular meshes to computational uniform grids in the geometric domain. However, it still does not work well when there is no diffeomorphism from the mechanic space to a uniform computational space.

\section{Problem Definition}

\begin{figure}[t]
\centering\vspace{-2ex}
\includegraphics[width=.9\linewidth]{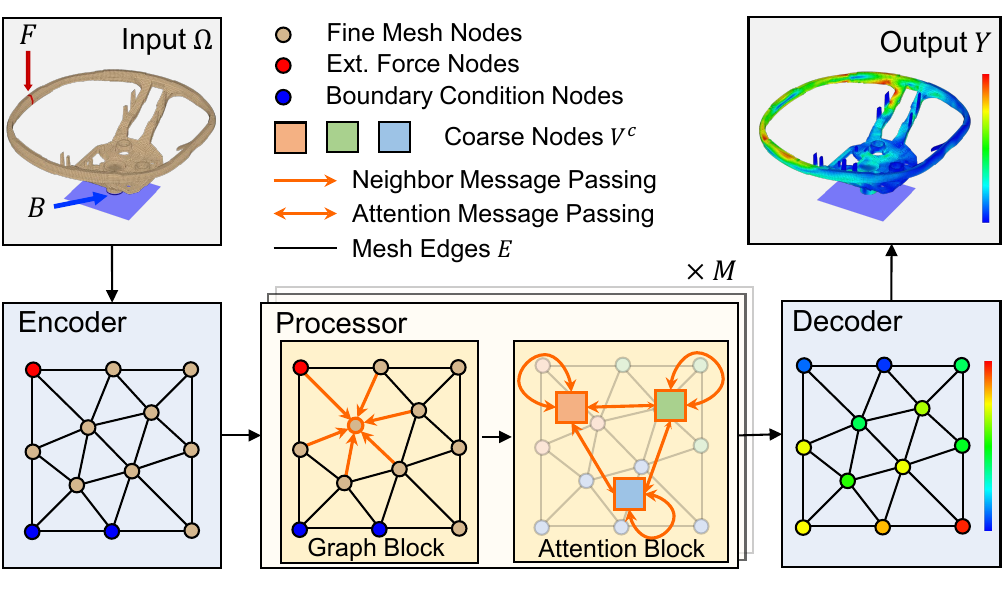}
\caption{Overall Framework}
\label{fig:framework}
\centering
\end{figure}

To learn mechanic simulation, we first model the mechanic system as a mesh graph as shown in Fig. \ref{fig:framework}. To this end, we can exploit a mesh generator that is nowadays widely provided by FEM tools, and discretize the mechanic system within a $d$-dimensional physical domain $\Omega \subseteq \mathbb{R}^d$ with $d = 2$ or 3 into  mesh structures. Depending on the simulation, these mesh structures consist of either surface elements or volume ones. Essentially, these discrete mesh structures, i.e., finite elements, approximate the geometrical shape or volume of the mechanic system. 

Next, we model discrete mesh structures by a mesh graph $G=(V, E)$. Each node $v_i\in V$ is with a $d$-dimensional coordinate $\mathbf{x}_i$. Denote $v_i$ and $v_j$ with $v_i\neq v_j \in V$ to be the endpoints of an edge, We then associate the edge with two displacement vectors $\vec{\mathbf{e}}_{ij} =\mathbf{x}_i-\mathbf{x}_j$ and $\vec{\mathbf{e}}_{ji} =\mathbf{x}_j-\mathbf{x}_i$. In this way, we do not maintain the absolute coordinates of graph nodes and instead the vectors of graph edges $E = \{ \vec{\mathbf{e}}_{ij}\}$. It makes sense because mechanic systems may be with their specific mesh node coordinate systems with various coordinate centers. Using such vectors mitigates the inconsistency issue across various coordinate systems (e.g., the centers of coordinate systems are located at various locations within mechanic systems).

After that, we model the boundary conditions, including the external force $F = \{ \mathbf{f}_i \}$ and fixing constraint $B = \{ \mathbf{b}_i \}$ over a subset of graph nodes $v_i$, as node features. That is, after  mesh generation, we reasonably assume that $F$ and $B$ are evenly applied to a subset of mesh nodes. For example, when the force $F = 100$ $\mathsf{newton}$s are applied to 20 graph nodes, each of such 20 nodes is assumed to be with 5 $\mathsf{newton}$s meanwhile other nodes with zero $\mathsf{newton}$. 

\begin{definition}[Complex Mechanic System Simulation]
Given a mechanic system $\Omega \subseteq \mathbb{R}^d$ modeled by a mesh graph $G=(V, E)$, external force $F$ and boundary fixing constraint $B$, we learn a regression model $R(\cdot)$ to generate the mechanic response $Y = \{\mathbf{y}_i\} \subseteq \mathbb{R}^{N \times p}$, i.e., $Y= R(G, F, B)$.
\end{definition}

In the problem, the prediction output $Y$ is the mechanic response $\mathbf{y}_i$  typically over every node $v_i$. For example in Fig. 1, we may predict a $p=2$ dimensional response $\mathbf{y}_i$ of effective stress and displacement.

\section{Methodology}

\subsection{Framework}

In Fig. \ref{fig:framework}, our solution follows an {Encoder-Processor-Decoder} framework. The Encoder first learns graph node and edge embedding vectors. Next, the Processor exploits two developed blocks, Graph Block (GBK) and Attention Block (ABK), to aggregate node embedding vectors via graph message passing. Finally, the Decoder generates the output $Y$ from the aggregated embedding vectors.

\subsubsection{Encoder} We perform the encoding operator on the input $\{G, F, B\}$ to generate node and edge embeddings, $\mathbf{v}_i \in V$ and $\mathbf{e}_{i, j}$, by the embedding function $\varepsilon^v$ and $\varepsilon^e$, respectively.

\begin{equation}\footnotesize
   \textstyle \mathbf{v}_i=\varepsilon^v\left(\mathbf{f}_i, \mathbf{b}_i\right), \quad \mathbf{e}_{i, j}=\varepsilon^e \left(\vec{\mathbf{e}}_{ij} ,||\vec{\mathbf{e}}_{ij} ||\right).
\end{equation}

In the equation above, $\mathbf{v}_i$ is the embedding vector of node $v_i$ regarding the external force $\mathbf{f}_i$ and boundary conditions $\mathbf{b}_i$ applied onto this node, and $\mathbf{e}_{i, j}$ is the embedding vector of the edge from node $i$ to node $j$ regarding the displacement vector $\vec{\mathbf{e}}_{ij} $ and its Euclidean distance $||\vec{\mathbf{e}}_{ij} ||$.

\subsubsection{Processor}
In Fig. \ref{fig:framework}, this stage involves a sequence of $M$ layers, each of which consists of a Graph Block (GBK) followed by an Attention Block (ABK). The GBK works within the input fine mesh graph $G$ to learn local neighbor interactions. Subsequently, the ABK performs on coarse graphs (the mesh graph coarsening algorithm will be given soon) to effectively capture the global structure of the mechanic system via Transformer. In this way, the GBK and ABK work together on the two-level fine and coarse mesh graphs to learn local and global representation. For the $l$-th layer with $1 \leq l \leq M-1$, we have

\begin{equation}\footnotesize
  \tilde{\mathbf{V}}^l, \mathbf{E}^{l+1} \leftarrow \operatorname{\textbf{GBK}}(\mathbf{V}^{l}, \mathbf{E}^{l}), \quad \mathbf{V}^{l+1} \leftarrow \operatorname{\textbf{ABK}}(\tilde{\mathbf{V}}^l, \mathbf{E}^{l+1})
\end{equation}

For the sake of readability, we do not specially mention the vectors on the $l$-th layer in the rest of the paper.

\subsubsection{Decoder}
This stage decodes node embeddings back into physical response output by a decoding function $\delta^v(\cdot)$. 
\begin{equation}\footnotesize
\mathbf{y}_i=\delta^v\left(\mathbf{v}_i\right).
\end{equation}

In the Encoder and Decoder above, we implement their associated functions such as $\varepsilon^v$, $\varepsilon^e$ and $\delta^v$ by MLPs.

\subsection{Graph Block}

This Graph Block (GBK) updates the node and edge embeddings, $\mathbf{v}_{i}$ and $\mathbf{e}_{i, j}$, via message passing within the fine graph $G$. Firstly, an edge embedding $\mathbf{e}_{i, j}$ is updated by the embeddings of the two endpoint nodes $\mathbf{v}_i$ and $\mathbf{v}_j$ of the edge to effectively learn their interactions. Subsequently, a node embedding $\mathbf{v}_i$ is updated by aggregating those updated edge embeddings $\mathbf{e}_{i,j}$ that connect to node $\mathbf{v}_i$.

 \begin{equation}\footnotesize
\begin{array}{cll}
\mathbf{e}_{i, j} &\leftarrow  \mathbf{e}_{i, j} \oplus f^E(\mathbf{e}_{i, j}, \mathbf{v}_i, \mathbf{v}_j) \\\\
\mathbf{v}_i &\leftarrow \mathbf{v}_i\oplus f^V(\mathbf{v}_i, \sum_{j \in \mathcal{N}_i} \mathbf{e}_{i, j})
\end{array}
\end{equation}

In the equation above, $f^E(\cdot)$ and $f^V(\cdot)$ denote the update functions regarding the edge and node embeddings, respectively, and $\mathcal{N}_i$ indicates the direct neighbors of node $v_i$. Here, unlike  traditional GNNs, we exploit {\emph{residual networks \cite{HeZRS16}}} to learn the changes between the original embeddings and updated ones. After that, we have the updated embeddings by the {addition operation} on the original embeddings and changes.

\subsection{Attention Block}

Unlike the GBK above, we develop the Attention Block (ABK) to learn global representation by a Transformer model. Recall that the Transformer requires computing intensive dot-product operations. Given an input graph $G$  with a large number of nodes, the Transformer on such a graph may suffer from substantial memory consumption and non-trivial computation overhead. To this end, in Fig. \ref{fig:attention}, we perform the ABK on a two-level mesh graph. Here, we generate a coarse graph $G^c$ on top of the input fine graph $G$. By using the \emph{simplified Louvain algorithm} \cite{blondel2008fast}, we can divide the nodes in the fine mesh graph $G$ into multiple groups and next map each group into an associated coarse mesh node in $G^c$. Here, the Louvain algorithm has been widely used for community detection with the efficient time complexity $O(N\cdot log N)$ where $N$ is the node count in the graph, and does not require the non-trivial efforts to pre-define or tune the number of communities. Given coarse mesh nodes, we connect two of them, if the fine graph $G$ contains at least one edge between the mapped fine nodes. After the coarse graph $G^c$ is generated, we now find that the node count in $G^c$ is much smaller than in $G$, and the Transformer on $G^c$ leads to higher efficiency.

Given the two-level mesh graph, we give the high-level workflow of the ABK as follows. Firstly, for every coarse mesh node in $G^c$ and the associated group of fine mesh nodes in $G$, the ABK \emph{aggregates} the fine mesh node embeddings in $G$ to the coarse mesh node embeddings in $G^c$. Next, the ABK employs the Laplacian position encoding to better represent the topology connectivity of the coarse mesh graph $G^c$. After that, the ABK exploits the Transformer on the coarse mesh graph $G^c$ to capture global respective information. Finally, the ABK disseminates the coarse mesh node embeddings from $G^c$ back to the fine node embeddings in $G$.

\subsubsection{Node Embedding Aggregation}
\begin{figure}[t]
\centering
\includegraphics[width=.9\linewidth]{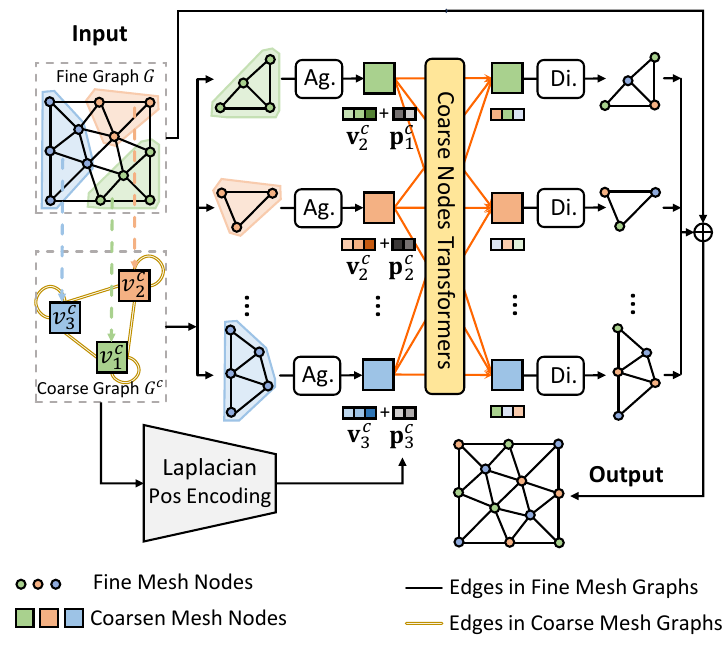}\vspace{-1ex}
\caption{Attention Block (Ag: Aggregate, Di: Disseminate).}\vspace{-2ex}
\label{fig:attention}
\centering
\end{figure}

As shown in Fig. \ref{fig:attention}, the ABK aggregates the node embeddings from the fine graph $G$ to the coarse graph $G^c$. Specifically, we assume that the coarse mesh graph $G^c$ contains $N^c$ nodes $\mathbf{v}^c_i$ with $1\leq i\leq N^c$. Denote $G(\mathbf{v}^c_i)$ to be the group of fine mesh nodes in $G$ that are mapped to the coarse node $\mathbf{v}^c_i$. Then, we define the following aggregation operation.

\begin{equation}\label{eq5}\footnotesize
\begin{array}{cll}
    \mathbf{v}^c_i& = & \frac{1}{|G(\mathbf{v}^c_i)|} \sum_{{v}_{k} \in G(\mathbf{v}^c_i)} \mathbf{v}_{k}
\end{array}
\end{equation}
In the equation above, we perform the aggregation operation by an average over the embeddings $ \mathbf{v}_{k}$ of fine nodes $\mathbf{v}_{k}$ in the group $G(\mathbf{v}^c_i)$. Such an average greatly reduces the number of learnable network parameters for higher efficiency. 

\subsubsection{Laplacian Position Encoding}

We exploit the Laplacian Position Encoding to better learn the geometry information of the coarse graph $G^c$. That is, for those nodes that are closer regarding their positions in the graph, the encoding leads to more similar positional features, and vice versa. Note that we have already used the relative coordinate displacement vectors $\vec{\mathbf{e}}_{ij} =\mathbf{x}_i-\mathbf{x}_j$, instead of absolute node coordinates $\mathbf{x}_i$. Such relative 
coordinate displacement vectors facilitate the encoding to meet the aforementioned goal. In addition, due to a smaller number of nodes in the coarse graph $G^c$ than the fine graph $G$, the ABK can achieve more efficient Laplacian Position Encoding on $G^c$.


The ABK follows the following step to compute the encoding. For the coarse graph $G^c$, we first need to compute a Laplacian matrix $\mathbf{L}^c \in \mathbb{R}^{N^c \times N^c}$ by $\mathbf{L}^c = \mathbf{D}^c - \mathbf{A}^c$, where $\mathbf{A}^c$ is the adjacency matrix of $G^c$ and $\mathbf{D}^c$ is the diagonal degree matrix regarding the node degree in $G^c$.

Next, we perform an eigen decomposition over the matrix $\mathbf{L}^c$. Denote $\{\mathbf{u}_1, \dots, \mathbf{u}_k\}$ to be the  eigenvectors with the top $k$ smallest non-zero eigenvalues. Such eigenvectors capture the most significant structural patterns of the graph $G^c$ at a global level. Here, each eigenvector $\mathbf{u}_i$ is a vector of length $N^c$, where $1 \leq i \leq k$. We can perform the decomposition by the widely used Lanczos algorithm with the  computational complexity $O(m \cdot k + k^2 \cdot N^c)$, where $m$ is the number of non-zero elements in the matrix  $\mathbf{L}^c$.

We then concatenate the elements of the $k$ eigenvectors to have the following  vector for each coarse node $\mathbf{v}^c_i$.

\begin{equation}\label{eq6}\footnotesize
    \mathbf{p}^c_i = [\mathbf{u}_1(i), \mathbf{u}_2(i), \dots, \mathbf{u}_k(i)] \subseteq \mathbb{R}^{k}
\end{equation}

Here, $\mathbf{u}_j(i)$ represents the $i$-th element of the $j$-th eigenvector. In spectral graph theory, the smallest eigenvalues (denoted as $\lambda$) indicate the most significant structure information  of the graph. By selecting eigenvectors associated with these smallest $k$ eigenvalues, the vector above can indicate the fundamental structure of the coarse graph by representing  essential graph connectivity while reducing the impact of high-frequency noise.

\subsubsection{Coarse Nodes Transformers}
\begin{figure}[t]
\centering
\begin{subfigure}[b]{0.35\linewidth}
    \centering
    \includegraphics[width=\textwidth]{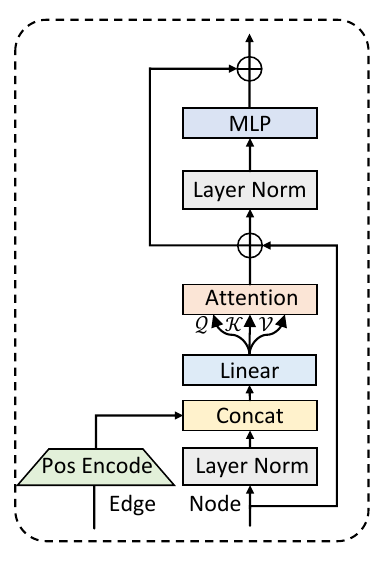}
    \caption{Transformer Layer.}
\end{subfigure}
\begin{subfigure}[b]{0.35\linewidth}
    \centering
    \includegraphics[width=\textwidth]{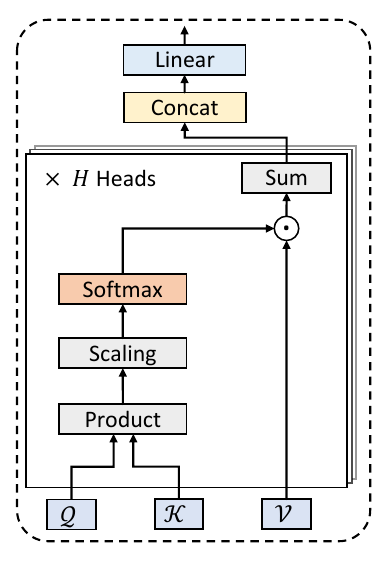}
    \caption{Self-Attention.}
\end{subfigure}
\vspace{-1ex}
\caption{The Transformer Layer and Multi-head Attention}
\label{fig:transformer}
\end{figure}

Until now, each coarse node is associated with two vectors $\mathbf{v}^c_i$ and $\mathbf{p}^c_i$ given by Eqs. (\ref{eq5} and \ref{eq6}). Next, denote $\mathbf{V}^c = \{\mathbf{v}^c_i\}$ 
and $\mathbf{P}^c = \{\mathbf{p}^c_i\}$
to be the entire vectors of all coarse mesh nodes in $G^c$. Then, the ABK employs Transformers on coarse mesh nodes to learn their spatial {dependencies}. As shown in Fig. \ref{fig:transformer}(a), the ABK first concatenates the layer-normalized node features $\mathbf{V}^c$ with the position-encoded features $\mathbf{P}^c$:

\begin{equation}\footnotesize
    \mathbf{Z} = \operatorname{Concat}[\operatorname{LayerNorm}(\mathbf{V}^c), \mathbf{P}^c]
\end{equation}

After that, the vector $\mathbf{Z}$ is then passed through a shared linear transformation to generate the query $\mathcal{Q}$, key $\mathcal{K}$, and value $\mathcal{V}$ vectors required for the attention mechanism:

\begin{equation}\footnotesize
    \mathcal{Q}, \mathcal{K}, \mathcal{V} = \operatorname{Linear}(\mathbf{Z})
\end{equation}

As shown in Fig. \ref{fig:transformer}(b), the attention mechanism computes attention scores by the dot product between queries and keys, scaled by the inverse square root of the key vectors' dimensionality $d_n$ to avoid overly large dot product values.

\begin{equation}\footnotesize
    {\mathbf{V}^c}^\prime = \operatorname{Softmax}\left(\frac{\mathcal{Q} \mathcal{K}^\top}{\sqrt{d_n}}\right) \mathcal{V}
\end{equation}

Subsequently, the ABK applies multi-head self-attention by replicating the attention mechanism by $H$ times independently (see Fig. \ref{fig:transformer}(b)). The outputs from these $H$ heads are concatenated and summed to generate a single vector, which is fed into the following layers.

\begin{equation}\footnotesize
    {\mathbf{V}^c}^{\prime\prime} = \operatorname{MLP}\left(\operatorname{LayerNorm}({\mathbf{V}^c}^\prime \oplus \mathbf{V}^c)\right)\oplus\left({\mathbf{V}^c}^\prime \oplus \mathbf{V}^c\right)
\end{equation}

After the operations above are applied, the updated coarse node embedding ${\mathbf{V}^c}^{\prime\prime}$ now involves the individual embedding and contextual relationships within the coarse graph. It enables the updated embedding to adaptively learn the complex patterns in the coarse mesh graph, providing a comprehensive representation of the global structure.

\subsubsection{Dissemination}
Finally, the ABK sends the updated embeddings of coarse nodes back to the original fine nodes.

\begin{equation}\footnotesize
    \mathbf{v}_i \leftarrow \{{\mathbf{v}^c_j}^{\prime\prime} \mid i \in G(\mathbf{v}^c_j)\} \oplus \mathbf{v}_i
\end{equation}

In this equation, if the fine mesh node $v_i$ belongs to the group $G(\mathbf{v}_i^c)$ of fine mesh nodes in $G$ that are mapped to the coarse node $\mathbf{v}_j^c$, the ABK then performs an {addition operation} over the embedding ${\mathbf{v}^c_j}^{\prime\prime}$ and its own embedding $\mathbf{v}_i$. Now, the updated embedding $\mathbf{v}_i$ incorporates both the global and local representation.

\subsection{Model Training}

To train the network, we first normalize the simulation input and output data using mean and variance. We employ an L2 loss function, $\mathcal{L} = \frac{1}{N} \sum_{i=1}^N \frac{1}{n_i} \sum_{j=1}^{n_i} \left(\mathbf{y}_{i,j}-\widehat{\mathbf{y}}_{i,j}\right)^2$, to measure the loss between prediction values $\widehat{\mathbf{y}}_{i,j}$ and ground truth $\mathbf{y}_{i,j}$, where $N$ denotes the number of datasets. $n_i$ represents the number of nodes in sample $i$. We use the Adam optimizer to train the neural network.
\section{Experiments}

\begin{table}[h]
\scriptsize 
\centering
\caption{Visualizations and Statistics of Four Used Datasets}
\setlength{\tabcolsep}{2pt} 
\begin{tabular}{p{1.5cm}p{1.2cm}p{2.1cm}p{1.2cm}p{1.8cm}} 
\toprule
\textbf{Dataset} & \textbf{Beam} & \textbf{Steering-Wheel} & \textbf{Elasticity}   & \textbf{CylinderFlow}  \\ \hline

Visualization       &\begin{minipage}[b]{0.16\columnwidth}
		\raisebox{-.5\height}{\includegraphics[width=\textwidth]{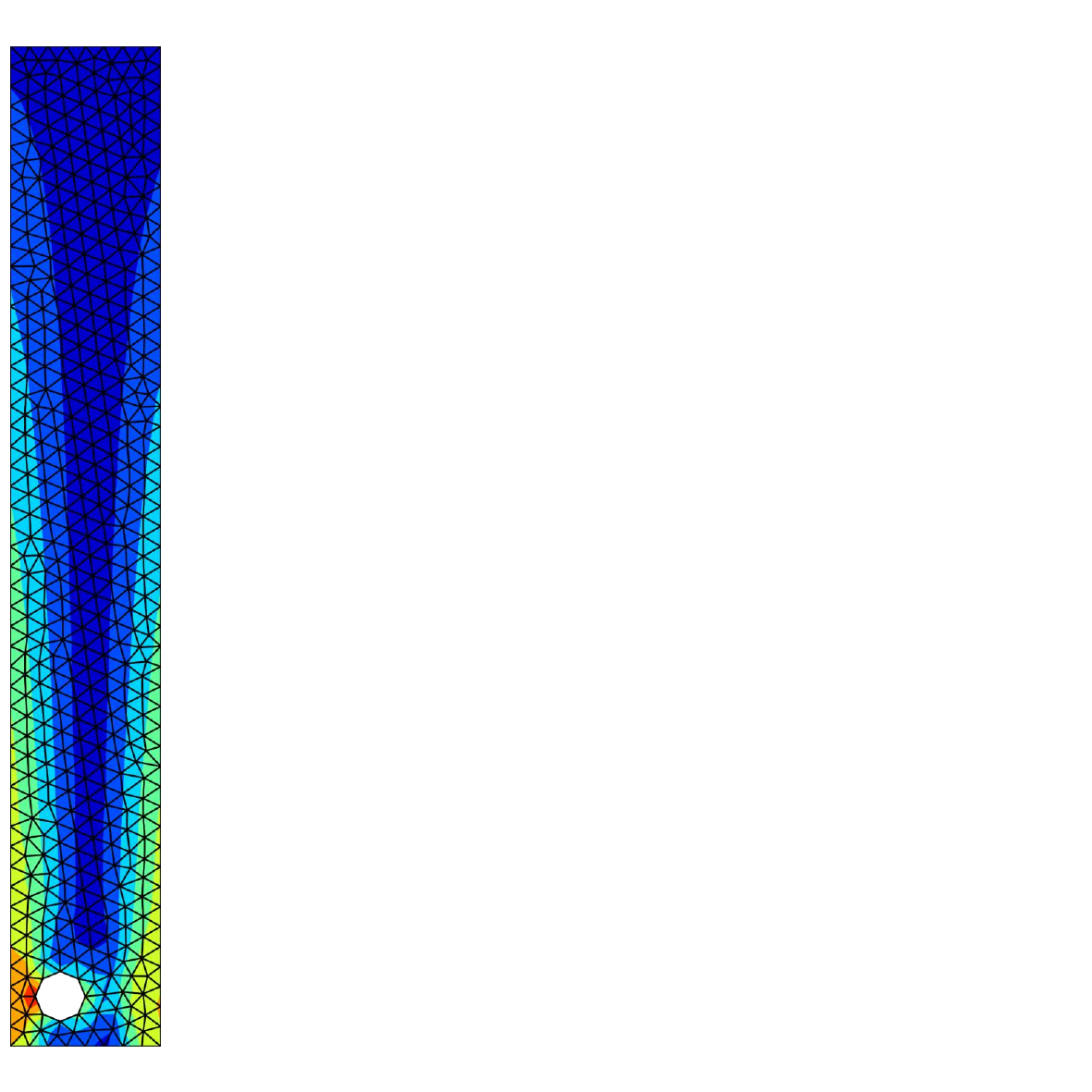}}
	\end{minipage}   &\begin{minipage}[b]{0.16\columnwidth}
		\raisebox{-.5\height}{\includegraphics[width=\textwidth]{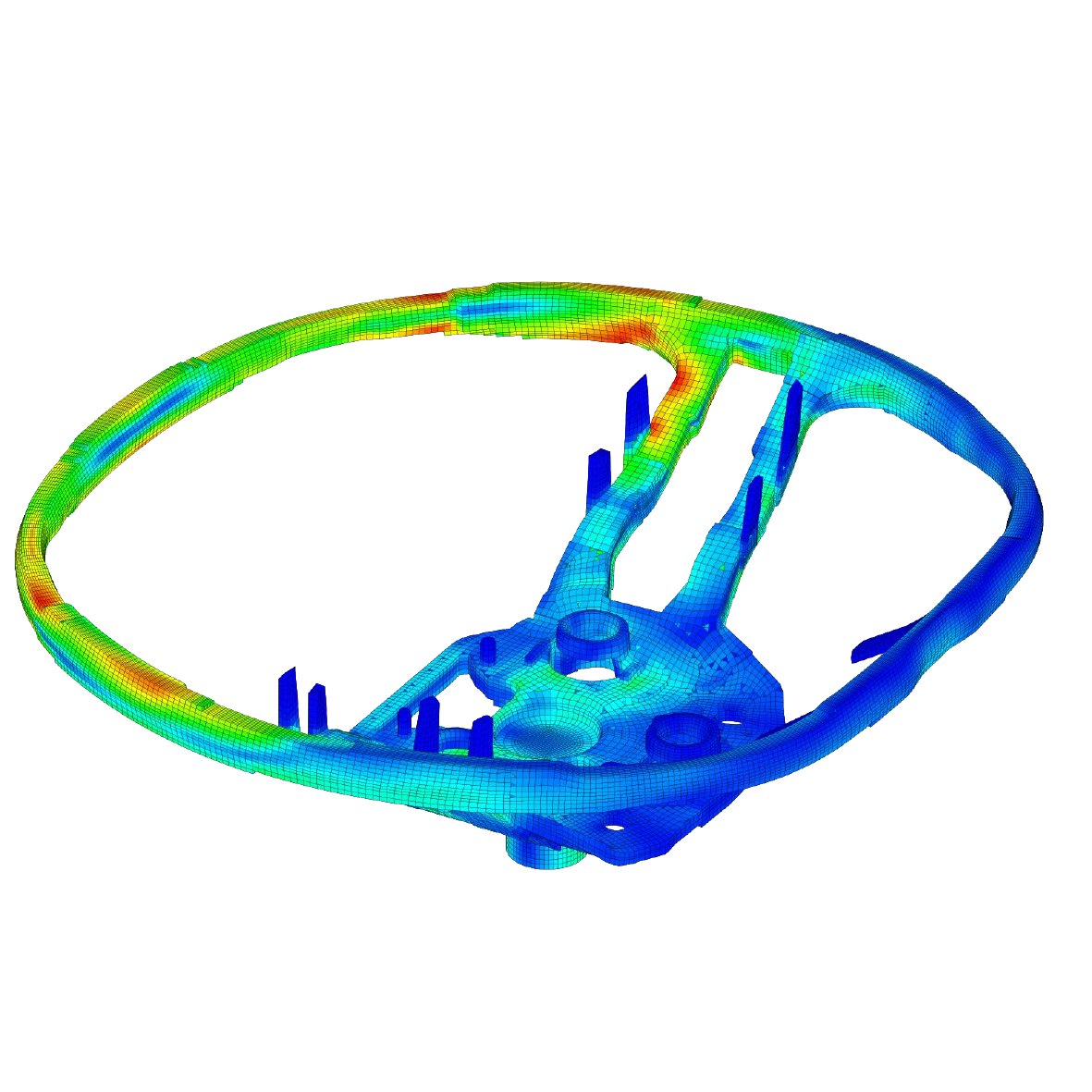}}
	\end{minipage}   &\begin{minipage}[b]{0.16\columnwidth}
		\raisebox{-.5\height}{\includegraphics[width=\textwidth]{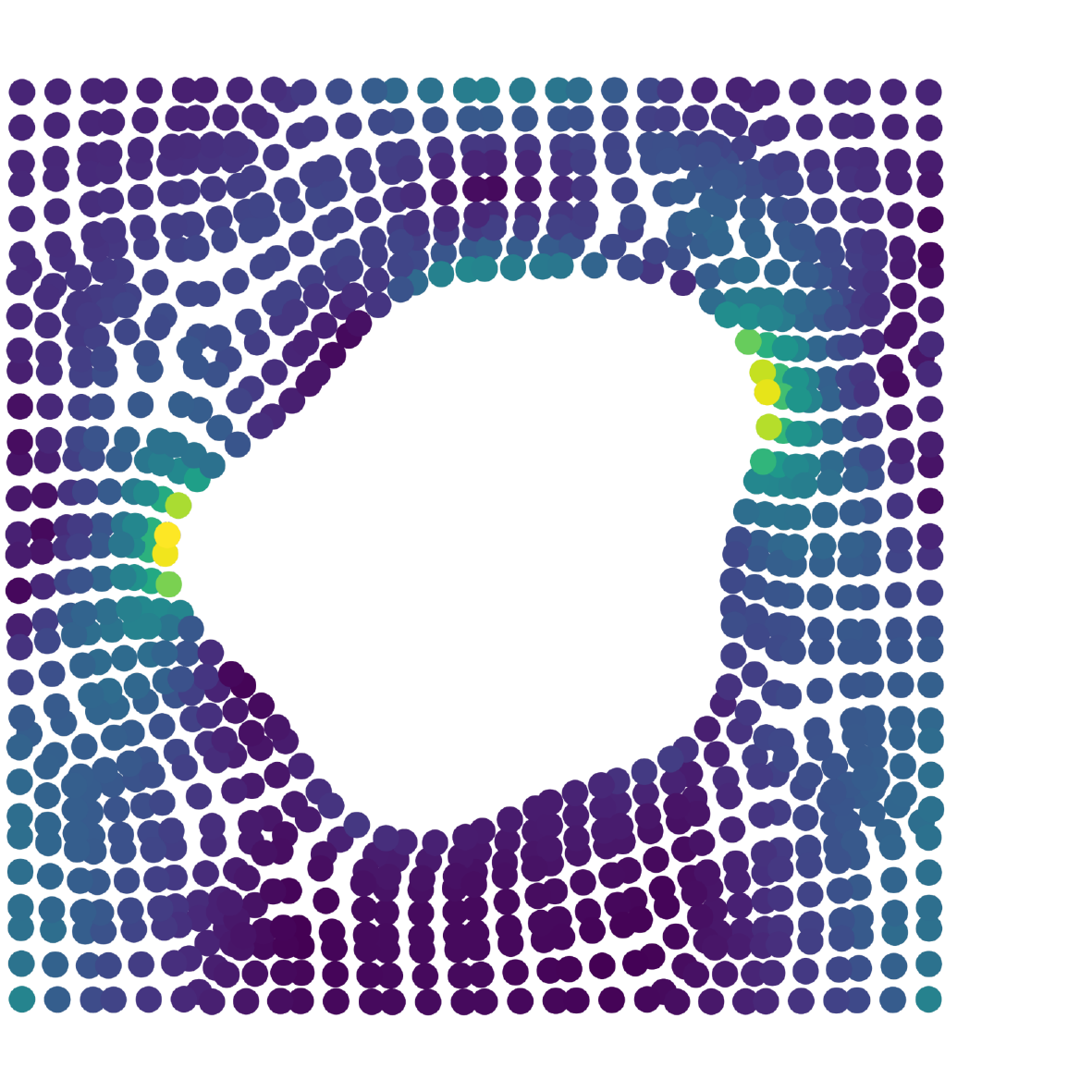}}
	\end{minipage}   &\begin{minipage}[b]{0.16\columnwidth}
		\raisebox{-.5\height}{\includegraphics[width=\textwidth]{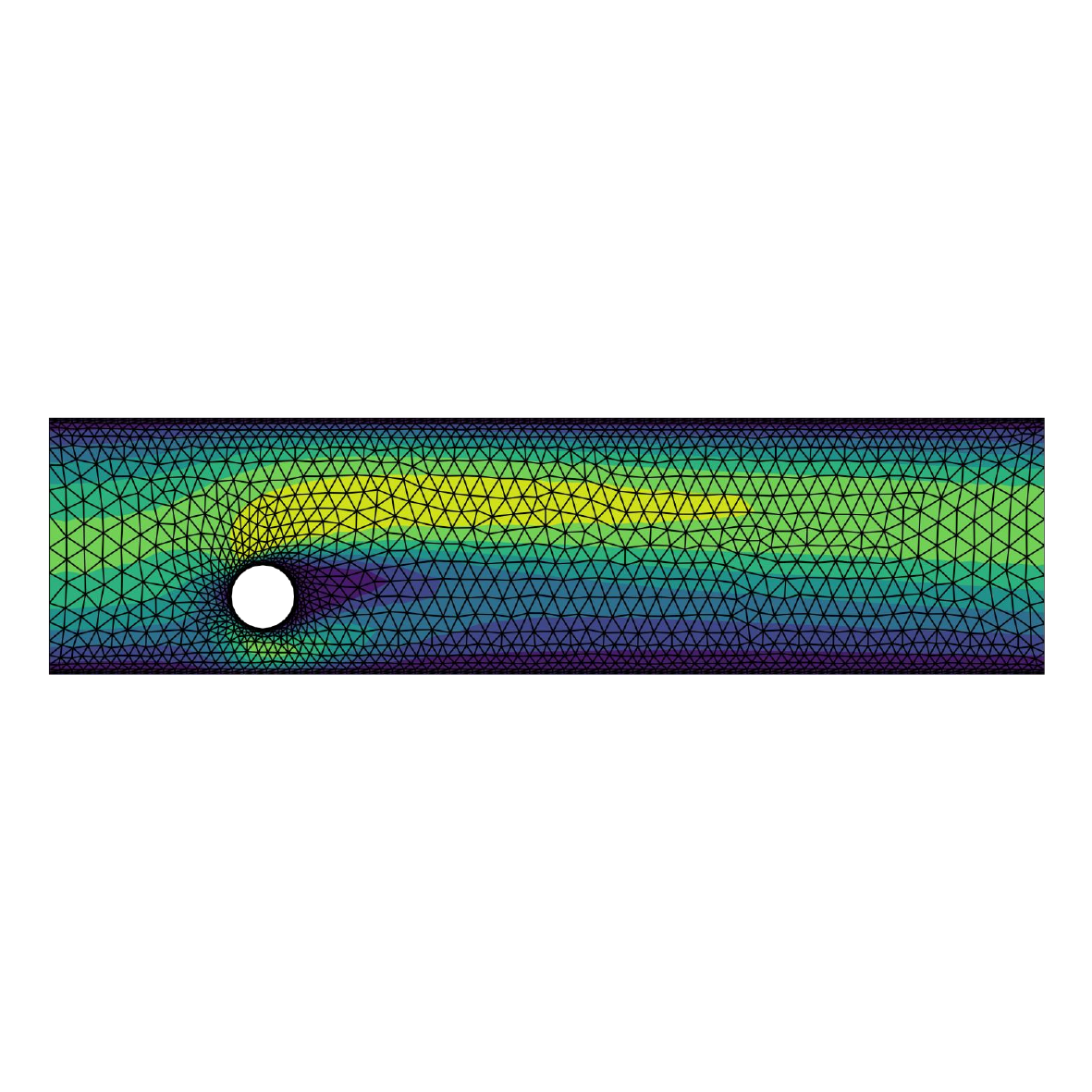}}
	\end{minipage}\\ \hline

\textbf{Samples}      & 555   &  239   &  2000 &  1200\\
\textbf{Avg. Nodes}      & 522.77   &  72061.01   &  972 &  1885.06\\
\textbf{Avg. Edges}      & 1444.32   &   200525.86   &  - &  5420.65\\
\textbf{FEM Solver}       & ABAQUS   & LS-DYNA   & PDE & COMSOL \\ 
\textbf{Phy. Response}           & Stress   & Stress   & Stress & Velocity, Pressure\\ 
\textbf{Mesh Type}      & 2D triangle   &  3D hexa-,tetra-hedral   & Point cloud &  2D triangle\\
\bottomrule
\end{tabular}
\centering\vspace{-2ex}
\label{table:datasets}
\end{table}

\subsection{Experimental Setting}

\subsubsection{Datasets}
We use one generated, one real and two open datasets for performance evaluation. 

\begin{itemize}
    \item \textbf{Beam}:  We use a popular FEM solver, ABAQUS \footnote{http://www.simulia.com/}, to generate this dataset on a 2D rectangular Cantilever Beam structure of the size $100 \times 15$ $mm^2$. A circle hole with a radius $r=2.5$ mm is within the beam structure. By varying the center position $(x, y)$ of the hole, we generate $111=3*37$ Beam objects, i.e., $(x, y) = (5+2.5*i, 5+2.5*j)$ with $i=0,1,2$ and $j=0,1,\dots,36$. We apply the external force $F$ with 300 $\mathsf{newton}$s with 5 various directions in the reverse direction of the $y$-axis at the end of the beam and boundary conditions at the other end. For an individual beam, we discretize its surface into 2D triangle grids by using the mesh generation tool provided by ABAQUS, and perform the FEM simulation to generate the mechanic response (as ground truth).

    
    \item \textbf{Steering-Wheel}: We use a real industrial data set provided by an automobile part supplier with 239 different steering wheel objects. Following the bending mechanic trial standard of automobile steering wheels,  expert engineers in the automobile company apply an external force $F$ with 700 $\mathsf{newton}$s in the reverse direction of the z-axis at the center of the steering wheel rim and meanwhile fix the wheel on a bottom plane. An industry-level FEM solver LS-DYNA\footnote{https://www.ansys.com/products/structures/ansys-ls-dyna} is used to simulate the torsion test and measure the resulting stress field in the steering wheel. Here, the generated mesh structure includes three mixed types of grids (hexa-, penta-, and tetra-dedral).
    
\item \textbf{Elasticity} \cite{10.5555/3648699.3649087}: This open dataset simulates the behavior of solid materials under various loading conditions. with input point clouds and output stress fields. We pre-process the dataset by the Delaunay triangulation method to ensure that the dataset can work for GNNs.
\item \textbf{CylinderFlow} \cite{PfaffFSB21}: This open 2D fluid mechanics dataset examines fluid dynamics around a cylindrical obstacle on vortex patterns, i.e., the Von Karman vortex street. Since this dataset contains 600-step time series data, we follow the work 
\emph{solver-in-the-loop} \cite{UmBFHT20}
to generate the rolling data from the initial time step $t=0$ to $t=1$, $t=250$ until 
the final $t=600$ time step, namely \textbf{+1}, \textbf{+250} and \textbf{Rollout}, respectively.
\end{itemize}

Given each dataset above, we choose 80\% samples for training, 10\% for validation and 10\% for testing. The details of the four datasets above refer to Table \ref{table:datasets}.
\subsubsection{Counterparts}

We compare our work against five recent works. For fairness, we have fully adopted the original parameter settings in the referred papers. 
\begin{itemize}
\item {\textbf{MeshGraphNets (MGN)}} \cite{PfaffFSB21} requires only one-level fine mesh graph and all message passing is performed on this flat graph. 
\item {\textbf{Multiscale MeshGraphNets (MS-MGN)}} \cite{abs-2210-00612} requires one fine mesh graph and a coarse one, leading to a two-level graph network. The message passing involves the down/up sampling across the two levels of graph networks. The approach uses the FEM solver to generate a coarse graph.
\item \textbf{{MultiscaleGNN} (MS-GNN)} \cite{abs-2205-02637} performs a multilevel hierarchical graph neural network, involving the message passing of down/up sampling across graph networks.
\item {\textbf{Eagle}} \cite{JannyBNDT023} employs node clustering, graph pooling and global attention to learn long-range dependencies between spatially distant graph nodes. Eagle converts \emph{absolute coordinates} into positional encodings by applying Sinusoidal Positional Encoding, which uses sine and cosine functions with varying frequencies for node position encoding \cite{VaswaniSPUJGKP17}.
\item \textbf{{Geo-FNO}} \cite{10.5555/3648699.3649087} is a geometry-aware discretization-convergent Fourier Neural Operator (FNO) framework that works on arbitrary geometries and a variety of input formats. The approach is designed for point clouds in solid materials and takes nodes' absolute coordinates as input to learn node embeddings.
\end{itemize}

\begin{figure}[ht]
\centering
\includegraphics[width=1.0\linewidth]{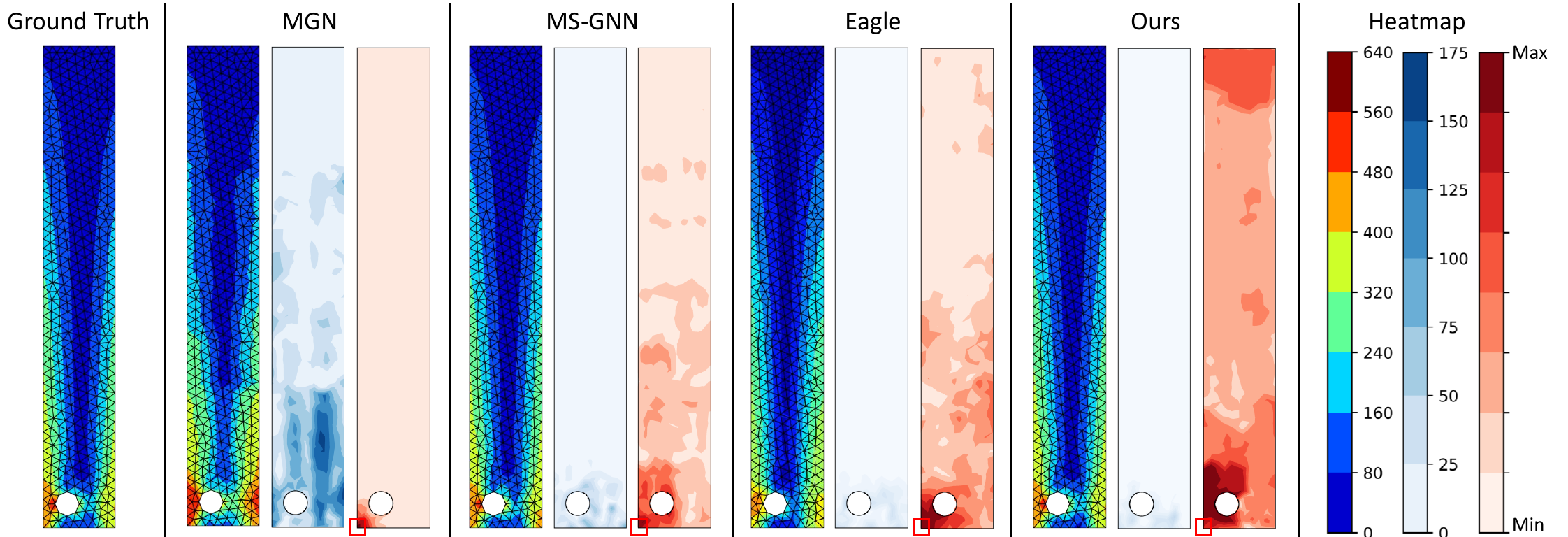}
\caption{Prediction and Gradient Visualization}\vspace{-2ex}
\label{fig:inter}
\centering
\end{figure}

\begin{figure*}[t]
\centering
\begin{subfigure}[b]{0.2\linewidth}
    \centering
    \includegraphics[width=\textwidth]{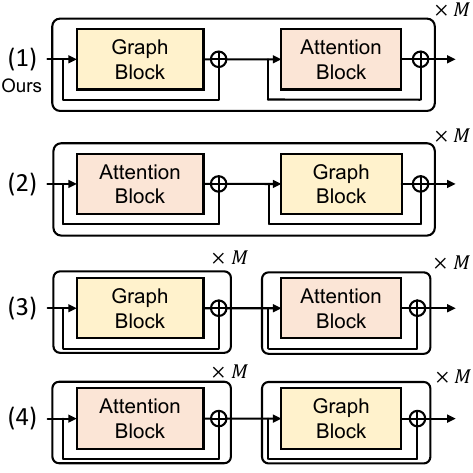}\vspace{-1ex}
    \caption{Net. Block Structures}
\end{subfigure}
\hfill
\begin{subfigure}[b]{0.25\linewidth}
    \centering
    \includegraphics[width=\textwidth]{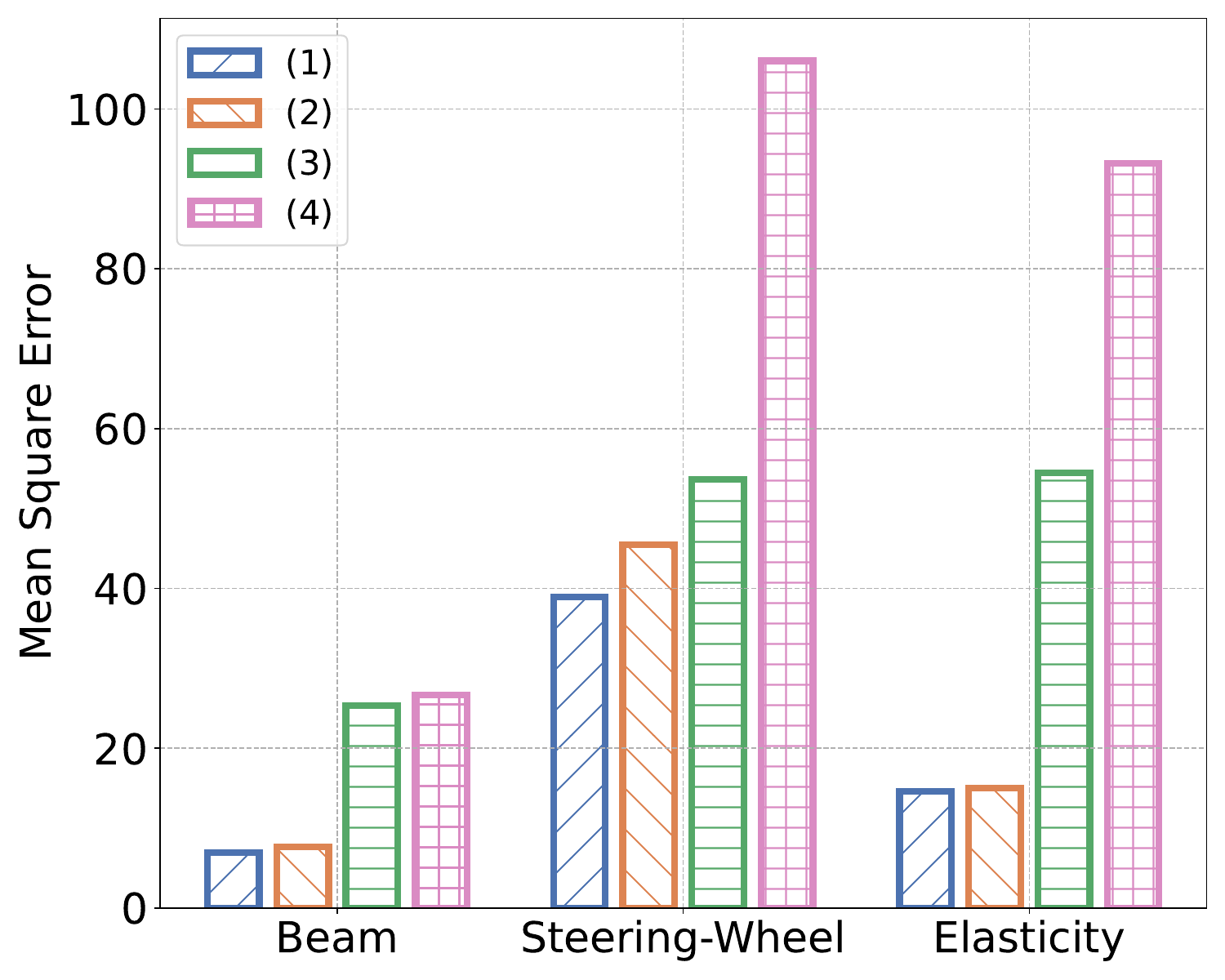}\vspace{-1ex}
    \caption{Effect of Net. Block Structures}
\end{subfigure}
\hfill
\begin{subfigure}[b]{0.2\linewidth}
    \centering
    \includegraphics[width=\textwidth]{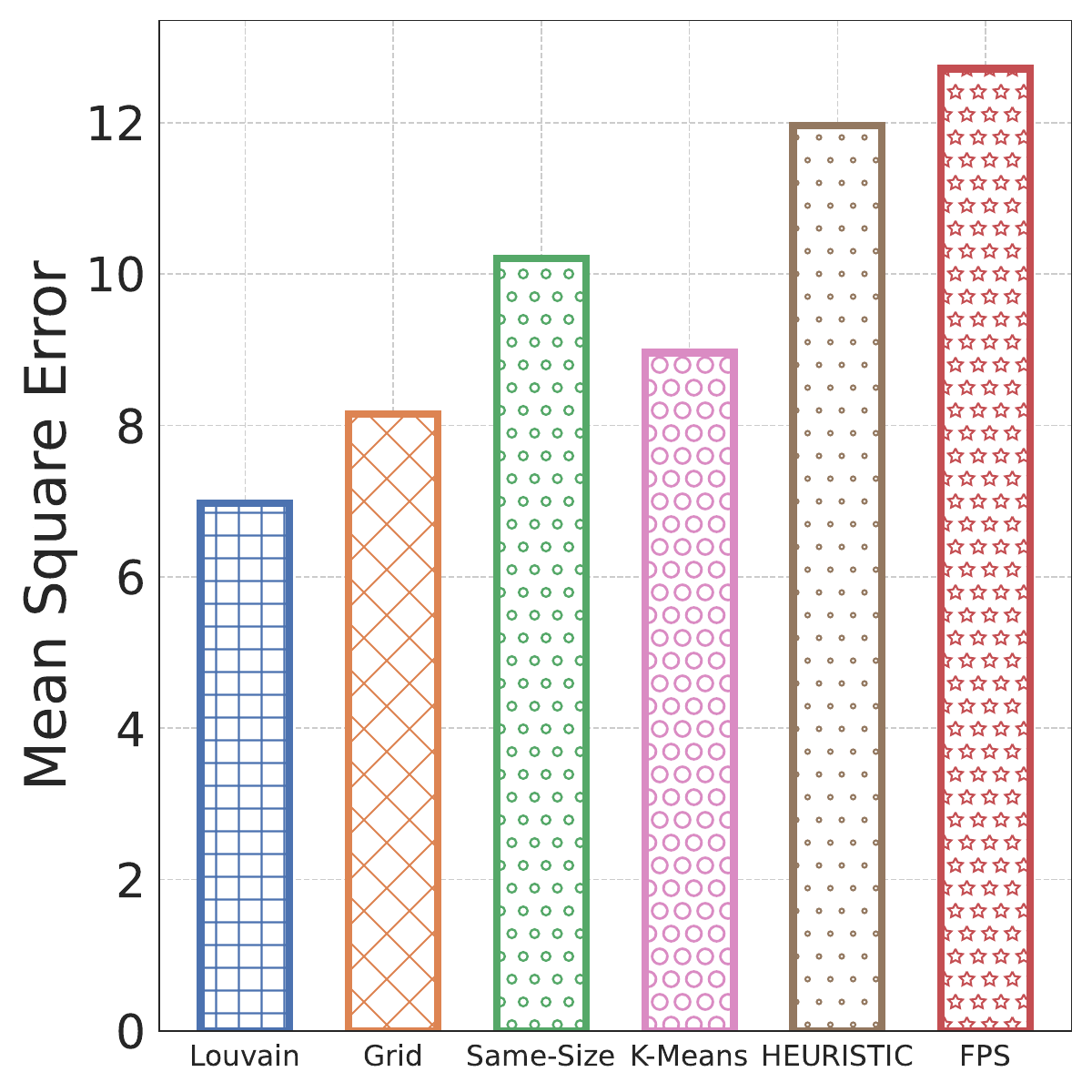}\vspace{-1ex}
    \caption{Coarsening Algorithms}
\end{subfigure}
\hfill
\begin{subfigure}[b]{0.2\linewidth}
    \centering
    \includegraphics[width=\textwidth]{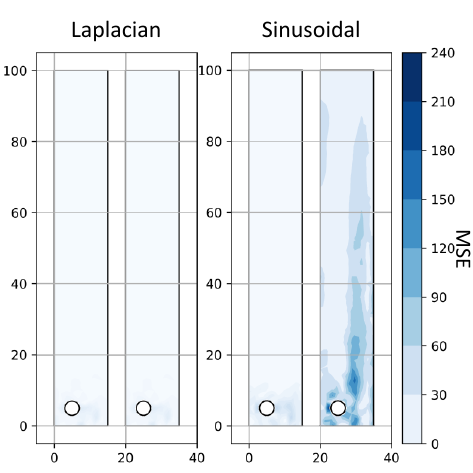}\vspace{-1ex}
    \caption{Pos. Encoding Schemes.}
\end{subfigure}
\vspace{-1ex}
\caption{Sensitivity Study}\vspace{-3ex}
\label{fig:exp}
\end{figure*}

\emph{Hyper-parameters Setting}: We use $M=7$ layers of the network blocks. The baseline methods use 15 iterations of message passing for the dataset. To ensure a fair comparison, our approach also utilizes similar parameters. We set the channel of hidden features $H$ as 128 and the number of hidden layers in MLPs as 2. Note that some methods involve multi-scale through message passing. For fairness, we set two scales of each method and approximate coarsen level of the graph, respectively. 


\subsubsection{Evaluation Metric} We measure the Mean Squared Error ($\mathrm{MSE}$) by $\frac{1}{N} \sum_{i=1}^N {\left(\frac{1}{n_i} \sum_{j=1}^{n_i} \left(\mathbf{y}_{i,j}-\widehat{\mathbf{y}}_{i,j}\right)^2\right)}$, where $N$ is the number of testing samples and $n_i$ is the number of nodes in sample $i$, and $\widehat{\mathbf{y}}_{i,j}$ (resp. $\mathbf{y}_{i,j}$) is the prediction (resp. ground truth) value of node $j$ in sample $i$.

We implement our prototype in Python 3.10 and models are written by PyTorch 1.13.0, and evaluate all the performance on the server equipped with an Intel(R) Xeon(R) W-2255 CPU @ 3.70GHz and NVIDIA 4090 GPU.

\footnotesize\vspace{-2ex}
\begin{table}[t]\scriptsize\vspace{-2ex}
\centering
\caption{Baseline result. Some works without prediction results (denoted by "N/A") due to the lack of coarse graphs in such datasets.}\label{table:baselines}\vspace{-2ex}
\begin{tabular}{p{1.1cm}p{0.7cm}p{0.7cm}p{0.7cm}p{0.4cm}p{0.4cm}p{0.4cm}p{0.6cm}}
\toprule
\multirow{2}{*}{\textbf{Methods}} & \multirow{2}{*}{\textbf{Beam}} & \multirow{2}{*}{\textbf{St-Wheel}} & \multirow{2}{*}{\textbf{Elasticity}} & \multicolumn{3}{c}{\textbf{CylinderFlow($\times E^{-04}$)}}    & \multirow{2}{*}{\textbf{Params.}}             \\
                                  &                                       &                                          &                       & \textbf{+1}                   & \textbf{+250}      & \textbf{Rollout}   &  \\
\hline
\textbf{MGN} & 5409.62 & 56.93 & 654.24 & 1.11 & \underline{6.16} & 14.57 & 2.33M\\
\textbf{MS-MGN} & 1364.83 & N/A & N/A & \textbf{\underline{0.95}} & 6.29 & \underline{13.63}  & 2.33M\\
\textbf{MS-GNN} & \underline{15.25} & 51.61 & 124.81 & 1.33 & 23.85 & 74.54  & 2.33M\\
\textbf{Eagle} & 17.18 & \underline{41.83} & \underline{15.88} & 2.57 & 17.65 & 71.97  & 10.32M\\
\textbf{Geo-FNO} &542.98 &424.14 & 17.88 & N/A & N/A & N/A  & 23.66M\\
\hline
\textbf{Ours} & \textbf{6.97} & \textbf{38.95} & \textbf{14.63} & 1.34 & \textbf{4.93} & \textbf{8.33} & 2.10M
\\ \bottomrule
\end{tabular}
\vspace{-2ex}
\end{table}
\normalsize
\subsection{Evaluation Result}
\subsubsection{Baseline Study} In Table \ref{table:baselines}, our work performs best on almost all datasets, for example, with 54.30\% lower errors than MS-GNN on the Beam dataset. Compared to the flat network model MGN, hierarchical models such as MS-MGN, MS-GNN, Eagle, and ours lead to lower errors. However, Geo-FNO does not work well with rather high errors due to irregular mesh structure in these datasets for complex mechanic simulation. In addition, from the result of the CylinderFlow time series data, our work performs best for the mid- and long-term prediction ($t=250$ and $t=600$ time steps) but not the very short-term prediction ($t=1$).

Besides, in the rightmost column of Table \ref{table:baselines}, we list the number of learnable network parameters for all models on the Beam dataset. Due to the adopted average operator in Eq. (\ref{eq5}), our work leads to 9.87\% fewer network parameters compared to MS-GNN. Here, the key insight is that our model outperforms the five competitors by the lowest MSE errors meanwhile with the fewest network parameters.

Fig. \ref{fig:inter} visualizes the ground truth of an example cantilever beam structure, prediction result of four models, including the predicted mechanic response (left), the difference between predicted response and ground truth (middle), and the overlaid gradient of each mesh node against the node at the lower left corner in the beam structure (right). Here, we follow the similar idea of Eagle \cite{JannyBNDT023} to compute the gradients. This figure clearly demonstrates that our result is the closest to the ground truth. In terms of the overlaid gradient, our work can precisely visualize the top area that is applied by the external force $F$. Such result is meaningful because the two works (MGN and MS-GNN) are inherently limited to very close neighborhood determined by the number of message passing, and the receptive field, which is represented as lower-left concentric square overlaid over the gradients. Yet, Eagle does not illustrate the top area applied by the external force. Instead, our model is not spatially limited and can pass messages across the entire scene (particularly those boundary areas) due to the interweaving ABK and GBK blocks on coarse mesh graphs.


\subsubsection{Study of Network Block Structure}  Fig. \ref{fig:exp}a gives four network block structures including ours and three alternatives. Here, in the first two structures (1-2), the GBK and ABK blocks (i.e., the residual networks) are both within each of the $M$ network layers, differing from the order of such two blocks, and the two rest structures (3-4) have the independent $M$-layer ABK (and GBK) blocks. 

As shown in Fig. \ref{fig:exp}b, our work, i.e., the (1) block structure, is with the lowest errors on the three datasets. Particularly, the first two structures (1-2) perform much better than the rest two (3-4). It is because the first two structures can ensure that each of the $M$ layer can learn both local and global node embeddings, and the $M$ layers work together to interweave such two embeddings for better representation. Moreover, the block structures with the GBK first order (1, 3) perform better than the ones with the ABK first (2, 4). The reason is that the first learned global representation by the ABK may otherwise obscure the local one by the GBK.



\subsubsection{Study of Coarsening Algorithms} In Fig. \ref{fig:exp}c, on the Beam dataset, we compare the used Louvain algorithm to generate coarse mesh node graphs against the following competitors.
\begin{itemize}
    \item Grid Sampling (Grid) \cite{abs-2205-02637}: By partitioning a multi-dimensional space into a set of grids, we choose those mesh nodes within a grid as a cluster.
    \item $k$-Means Sampling: This method applies the original $k$-Means clustering algorithm, which partitions the data into $k$ clusters by minimizing the variance within each cluster. Each data point is assigned to the nearest cluster center, and these centers are iteratively updated.
    \item Same-Size-$k$-Means Sampling (Same-Size) \cite{GanganathCT14}: An adaptation of the $k$-Means clustering algorithm that, in addition to minimizing variance, also ensures each cluster has approximately the same number of elements.
    \item Heuristic Uniform Sampling (HEURISTIC): We first randomly pick a single seed node and choose its $k$-hop nearest nodes as a cluster. We repeat this step among the remaining nodes until all nodes are clustered.
    \item Farthest Point Sampling (FPS)\cite{QiYSG17}: By randomly choosing an initial point, FPS iteratively selects the point that is farthest from the already selected points, until the desired number of points is reached. FPS ensures that the selected points are well-distributed to capture the essential characteristics of the dataset.
\end{itemize}

For fairness, we expect that all these algorithms can generate coarse graphs with the equal number of coarse mesh nodes. To this end, since the Louvain algorithm does not pre-specify the number of communities (or clusters), we first apply the Louvain algorithm to generate coarse mesh node graphs (with the node count 14 on average) on the input fine graphs (with 523 mesh nodes on average). Next, by using the count of such coarse nodes as input, the five rest algorithms then generate the associated coarse node graphs.

From this figure, we find that the Louvain algorithm leads to the lowest error. This is because the Louvain algorithm divides input graph nodes into multiple groups mainly depending upon the graph topology connectivity. Other algorithms, such as $k$-means and its variant Same-Size-$k$-means, mainly exploit node coordinates to compute node distance, and thus may not capture the graph topology, despite their widespread use in point clouds.



\subsubsection{Study of Position Encoding Schemes}
To demonstrate the superiority of the Laplacian position encoding, we compare our Laplacian scheme against the Sinusoidal scheme \cite{VaswaniSPUJGKP17}, which is adopted by the recent work Eagle \cite{JannyBNDT023}. To be consistent with the work Eagle, we use the Sinusoidal scheme to encode the absolute position coordinates of mesh nodes by applying sine and cosine functions with varying frequencies. To evaluate the performance, we purposely change the centers of coordinate systems in our testing data, by horizontally moving its original coordinate center by 20 $mm$. In this way, we can study how the position encoding scheme can adapt such a change.

Fig. \ref{fig:exp}d plots the difference between the prediction result and ground truth. For each encoding scheme, we have two bars: the left bar is the original result before the change of coordinate centers and the right one is the result after the change. As shown in this figure, when compared to the Sinusoidal scheme, in the two bars, the Laplacian scheme both leads to the lower errors. Such result indicates the adopted Laplacian scheme can best represent mesh nodes, no matter coordinate centers change or not.


\section{Conclusion}

In this paper, we propose a novel two-level mesh graph network to represent complex mechanic interaction between simulation objects, external force and boundary constraints. With the developed Graph Block and Attention Block, we design the $M$-layer network to interweave such two blocks for better local and global representation. Evaluation on three synthetic and one real datasets demonstrates that our work outperforms the state-of-the-art by better effectiveness and efficiency, e.g, with $54.30\%$ lower prediction errors and 9.87\% fewer network parameters on the Beam dataset. As future work, we are interested in the unified model to learn mesh generation and mechanic simulation and also plan to extend our model for more general simulation.

\newpage

\end{document}